\documentclass[journal]{IEEEtran}
\usepackage{cite}
\usepackage{amsmath}
\usepackage{graphicx}
\usepackage{subfig}
\usepackage{multirow}
\usepackage{hyperref}
\hypersetup{hypertex=true,
            colorlinks=true,
            linkcolor=blue,
            anchorcolor=blue,
            citecolor=blue}

\usepackage{booktabs}
\usepackage{stfloats}

\begin{document}

\title{Constructing a High Temporal Resolution Global Lakes Dataset via Swin-Unet with Applications to Area Prediction
}

\author{Yutian~Han,
        Baoxiang~Huang,~\IEEEmembership{Member,~IEEE,}
        He~Gao

\thanks{Yutian~Han is with Qingdao University, Qingdao, 266071, China (e-mail: 2021204134@qdu.edu.cn).}
\thanks{Baoxiang Huang is with the College of Computer Science and Technology, Qingdao University, Qingdao 266071, China. And also with the Laboratory for Regional Oceanography and Numerical Modeling, Laoshan Laboratory, Qingdao,266100, China.(e-mail: hbx3726@163.com).}
\thanks{Corresponding author: He~Gao is with the College of Computer Science and Technology, Qingdao University, Qingdao 266071, China(e-mail: ghsticker@163.com).}}

\maketitle

\begin{abstract}

Lakes provide a wide range of valuable ecosystem services, such as water supply, biodiversity habitats, and carbon sequestration. However, lakes are increasingly threatened by climate change and human activities. Therefore, continuous global monitoring of lake dynamics is crucial, but remains challenging on a large scale. The recently developed Global Lakes Area Database (GLAKES) has mapped over 3.4 million lakes worldwide, but it only provides data at decadal intervals, which may be insufficient to capture rapid or short-term changes.This paper introduces an expanded lake database, GLAKES-Additional, which offers biennial delineations and area measurements for 152,567 lakes globally from 1990 to 2021. We employed the Swin-Unet model, replacing traditional convolution operations, to effectively address the challenges posed by the receptive field requirements of high spatial resolution satellite imagery. The increased biennial time resolution helps to quantitatively attribute lake area changes to climatic and hydrological drivers, such as precipitation and temperature changes.For predicting lake area changes, we used a Long Short-Term Memory (LSTM) neural network and an extended time series dataset for preliminary modeling. Under climate and land use scenarios, our model achieved an RMSE of $0.317 km^2$ in predicting future lake area changes.
\end{abstract}


\section{Introduction}
Lakes are vital water bodies on the Earth's surface that provide crucial environmental\cite{function_lake_environmental}, economic, and cultural services\cite{function_lake_cultural}. However, climate change and human activities are disrupting their hydrological conditions and water balances\cite{affected_by_climate1, affected_by_climate2,affected_by_climate3}. Elevated temperatures, altered precipitation patterns, and heightened evapotranspiration rates, intrinsically linked to climate change, coupled with unsustainable human water extraction practices, are exerting adverse impacts on the water budgets of lakes\cite{affected_by_evaporation, affected_by_temperature,affected_by_extraction}. Declining lake levels threaten biodiversity\cite{threat_biodiversity} and ecosystem services like water supply\cite{threat_water_supply}, flood regulation\cite{threat_flood_regulation}, fisheries\cite{threat_fisher}, and recreation\cite{threat_recreation}. Continuous global monitoring of lake dynamics is crucial to understand the impacts of environmental change on these freshwater ecosystems, but databases depicting worldwide variability in lake areas over time are scarce.

Existing lake databases have provided valuable standardized data on lake conditions, but are limited in spatial coverage and resolution. Natural Lake Dataset in China encompasses over 40,000 natural lakes in China during 1985-2020, but covers only China without a global extent\cite{NLD_China}. ILEC World Lake Database focuses on providing environmental and socioeconomic data for over 500 significant lakes globally, but has limited spatial representativeness centered on human-use values\cite{ILEC_World_Lake_Database}. Recent studies have characterized long-term changes in lakes at the global scale using remote sensing data and deep neural networks. \cite{GLAKES} mapped area variations of 3.4 million lakes over the past four decades at a 10-year interval, producing the most spatially comprehensive database of global lake dynamics to date called GLAKES. However, the relatively coarse temporal resolution limits further elucidation of detailed lake variations.
While constructing a comprehensive dataset is crucial, it is imperative to develop prognostic models focusing on salient features such as lake areas for anticipating future dynamics. Predicting changes in lake water levels is a difficult endeavor, since lake level variations are impacted by numerous elements, including weather, hydrology, and geology. While traditional mathematical models like physically-based\cite{	predict_math_physical}, probabilistic\cite{predict_math_probabilistic} and system dynamics\cite{predict_math_sd} models have been extensively used for forecasting lake water levels, machine learning techniques have emerged as a promising alternative approach in recent studies\cite{predict_ML1,predict_ML2,predict_ML3}. Machine learning methods like artificial neural networks can learn complex nonlinear relationships between predictors like climate variables and lake levels from historical data\cite{predict_DL1,predict_DL2, predict_DL3}. They offer advantages of requiring less detailed domain expertise of physical processes and can continuously improve predictive performance by adapting to new data.

\begin{figure*}[ht]
\centering
\includegraphics[width=1\textwidth]{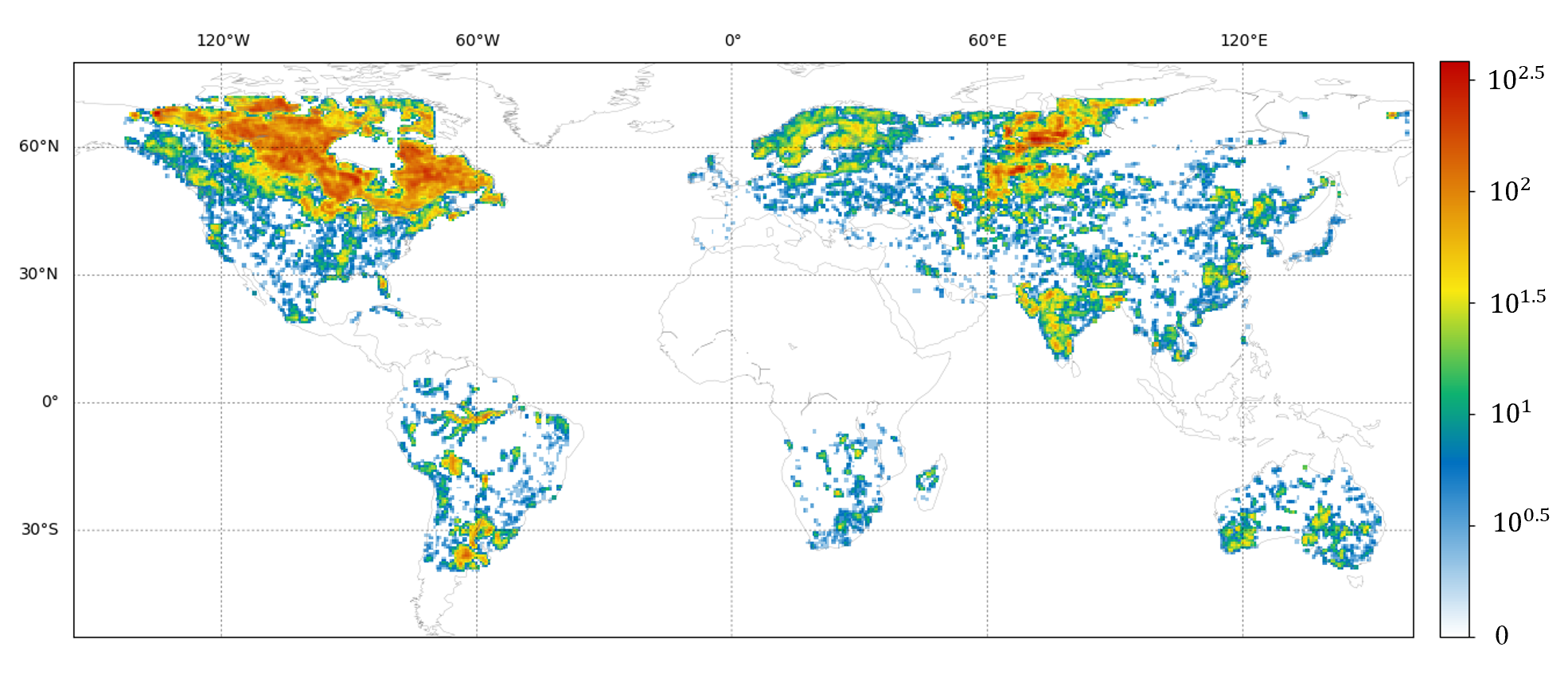}
\caption{Global 0.5°x0.5° gridded map depicting the lake count per grid cell, derived from the GLAKES-Additional dataset obtained in this study.}
\label{Fig1}
\end{figure*}

Here we present a new global lake dynamics database, GLAKES-Additional, that provides biennial changes from 1990 to 2021 for a subset of lakes contained in the GLAKES database. Through applying Swin-Unet on global satellite images, we delineated lake boundaries and extracted areas every two years over 16 epochs, generating a database encompassing over 152,567 lakes greater than 0.5 $km^2$ distributed across all continents except Antarctica. The global distribution of the lakes obtained is illustrated in Fig. \ref{Fig1}. As a complement and enhancement to the decadal GLAKES database, GLAKES-Additional allows cross-referencing with GLAKES using a common Lake\_id field. The doubled temporal resolution offers improved quantification of lake variations and enables more accurate attribution of associated drivers. Open access to this dataset advances the capability to monitor lake changes globally in the context of accelerating anthropogenic stressors and climate change. By tracking fluctuations in inland water bodies worldwide, the database contributes to an enhanced scientific understanding of lake ecosystem dynamics, supporting wiser decision-making and policies aimed at sustaining these critical freshwater resources on which humanity and biodiversity rely.
And our GLAKES-Additional dataset provides near-annual temporal resolution measurements of lake areas across the globe, enabling lake area prediction at a global extent. We integrated the time series of lake area from GLAKES-Additional with meteorological data including precipitation , temperature, and vapor pressure to construct joint sequences at at a 2-year interval. These multi-variable time series were used to train LSTM to model lake area changes under climatic influences and predict future variations. Preliminary results show promising performance of the data-driven LSTM models in predicting future lake area variations, achieving an RMSE of $0.317 km^2$ by learning from past area fluctuation patterns under varying climatic conditions.

The rest of the paper is organized as follows. The dataset 
and the preprocessing procedure are discussed in Section \ref{Data description}. 
The architecture of the proposed Swin-Unet is described in 
Section III. In Section IV, the results of eddy identification by 
different models are compared. Finally, Section V concludes this 
paper.

\section{Data description}
\label{Data description}

\subsection{JRC Monthly Water History}
The Monthly Water History dataset utilized in this study is part of the Global Surface Water product generated by the European Commission's Joint Research Centre\cite{GSW}. This dataset was derived from analysis of over 4.7 million Landsat satellite images spanning March 1984 to December 2021. Each 30-meter pixel in the images was classified as either water or non-water through an expert machine learning algorithm developed by the data producers. The per-pixel classification results were then aggregated into 454 monthly maps from 1984 to 2021, providing a global quantification of surface water distribution changes over time. the Monthly Water History dataset provides key insights into environmental changes and can support various applications including water resource management, climate and hydrological modelling, biodiversity monitoring, and food security assessments.
\subsection{Global Surface Water Occurrence}
The Global Surface Water Occurrence dataset (GSWO) also originate from the Global Surface Water project led by the European Commission's Joint Research Centre\cite{GSW}. The GSWO encapsulates the overall dynamics of surface water from 1984 to 2021 by depicting the frequency of observed surface water during this period. Water occurrence is quantified as the percentage of available satellite observations that identified surface water for each location, normalized to accommodate variations in observational density over time. This normalization enables consistent characterization of surface water dynamics across the globe throughout the multi-decadal period. The GSWO dataset was derived using a consistent preprocessing approach as the two-year cyclic water occurrence frequency dataset mentioned above. The GSWO product and corresponding manual annotations were utilized to train a Swin-UNet model for automated lake extraction. The trained Swin-UNet model was then applied to the two-year cyclic water occurrence frequency dataset to delineate lake boundaries at each time step. This allowed for efficient extraction of lake areas from the satellite imagery in a consistent and reproducible manner.
\begin{figure*}[ht]
\centering
\includegraphics[width=1\textwidth]{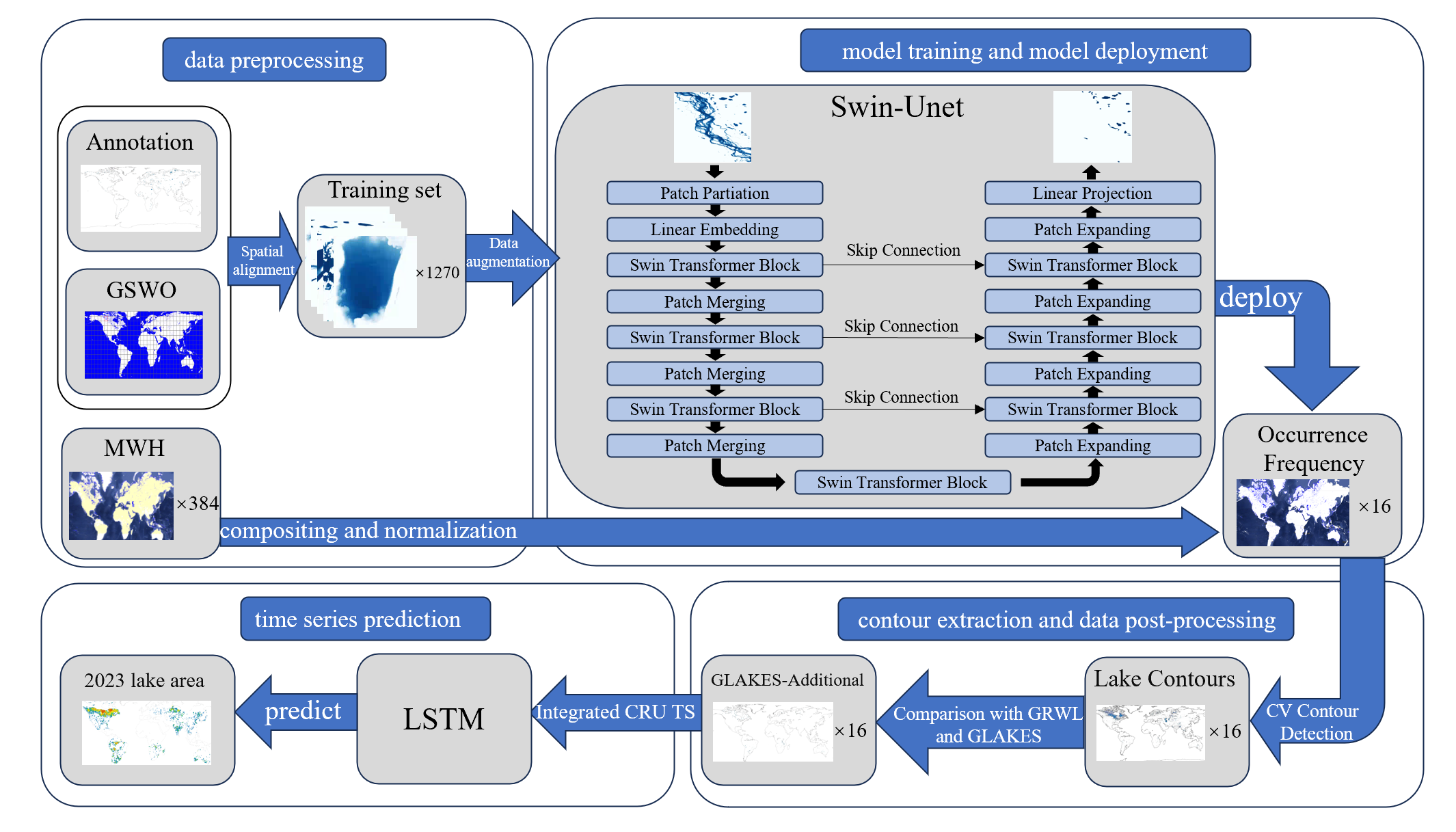}
\caption{The workflow for constructing a high-resolution global lakes dataset using Swin-UNet, and predicting lake area changes utilizing a stacked LSTM network driven by meteorological data.}
\label{fig1}
\end{figure*}
    
\subsection{GSHHS}
The Global Self-consistent Hierarchical High-resolution Shoreline (GSHHS) is a comprehensive global coastline dataset compiled from two public domain databases\cite{GSHHS}. With five levels of resolution - full, high, intermediate, low and crude, it contains polygons enclosing the continents and major islands. GSHHS has undergone extensive processing to remove inconsistencies and errors. The shorelines are hierarchically constructed from closed polygons, enabling efficient spatial searches and analysis. Its wide utility includes geographic masking, data selection and studying statistical properties of coasts.

In our work, we utilize the GSHHS low-resolution coastlines as a land mask to exclude ocean areas. The low resolution sufficiently meets our masking needs while greatly reducing data volume and processing complexity. By adopting this standardized public domain shoreline resource, we improve reproducibility and streamline our workflow.
\subsection{GRWL}
The Global River Widths from Landsat (GRWL) Database was developed to characterize the global coverage of rivers and streams\cite{GRWL}. It represents the first global compilation of river planform geometry under constant discharge conditions. The database contains width measurements for over 2 million km of rivers $\geq$ 30 m wide, acquired from Landsat imagery during months of mean discharge. In situ validation indicated that the GRWL database provides optimal accuracy for rivers wider than 90 meters. By quantifying geometry for global river networks, the GRWL Database provides an invaluable data source to investigate fluvial processes and interactions with climate, hydrology, and biogeochemistry. Details on how we utilized the GRWL Database in our study will be introduced in the following sections.
\subsection{CRU TS, v4.07}
The Climatic Research Unit gridded Time Series (CRU TS) dataset provides monthly land surface climate data from 1901 onwards with 0.5° resolution, covering global land areas excluding Antarctica\cite{CRU_TS}. It incorporates quality-controlled observations from extensive station networks and interpolates them into gridded values for ten key climate variables using angular-distance weighting. The datasets undergo regular updates and rigorous validation. A defining aspect of CRU TS is its complete spatial coverage through climatological infilling, facilitating diverse climate applications. For this research, CRU TS supplied crucial historical climate variables like precipitation and temperature as inputs to model and analyze lake area changes across the study period.

\section{Methodology}
The overall workflow of the methodology employed in this study is illustrated in Fig. \ref{fig1}. Initially, Spatially align GSWO and annotation data to generate training samples, and derive global water occurrence frequency over 16 time steps from MWH data. Subsequently, two distinct Swin-Unet models are trained on the prepared dataset, and then applied to the 16-step global water frequency dataset for pixel classification. Following this, a contour detection technique in OpenCV is utilized to generate lake contours, which are further refined to produce GLAKES-additional. In the final step, the area information from GLAKES-additional is integrated with CRU TS data to construct a temporal dataset, and LSTM is employed to forecast the lake area.

\subsection{Data Processing}
To generate a training dataset, we utilized the open-source lake annotations provided by \cite{GLAKES}. These annotations are available in vector format, employing the same WGS84 geographic coordinate system as the GSWO. We extracted relevant data from the GSWO dataset using predefined areas of interest in vector format and performed spatial alignment. Subsequently, we rasterized the corresponding lake annotations within these AOIs to generate training labels. This process yielded 802 samples from non-flooded regions and 468 samples from flood-affected areas. 

Given the relatively small size of our dataset, we employed image augmentation techniques to simultaneously process and augment both the data and labels. First, we randomly padded each sample to ensure its height and width exceeded 512 pixels, using zero padding to represent areas where the water occurrence frequency was zero, indicating non-lake regions. Subsequently, we normalized each image by subtracting its mean and dividing by its standard deviation, a common practice in machine learning to improve numerical stability and convergence during model training. Next, we randomly cropped the images to a size of $ 512 \times 512$ pixels. Finally, we further augmented the dataset by applying random horizontal and vertical flipping operations to help models learn rotation-invariant features.

The Monthly Water History data was acquired using Google Earth Engine (GEE), a cloud-based platform enabling planetary-scale geospatial analysis by combining satellite imagery catalogs and geospatial datasets. GEE allows developing JavaScript code using their web-based editor to define computational workflows which are then executed on Google's infrastructure. To optimize bandwidth, we consolidated 24 monthly images over two years into a single two-year composite image on the server-side. Only the final merged result was downloaded, greatly reducing data transfer requirements. During the compositing process, pixel values were normalized by dividing them by the number of valid observations over the two-year period. Due to some regions having no valid observations during the two-year period, a GEE-derived mask was applied to omit those locations. Prior to download, the resultant image was manually subset to areas of interest to further minimize bandwidth needs. We ultimately obtained a novel dataset similar to GSWO, where each pixel value represents the probability of water occurrence at that location, providing global water occurrence frequency over 16 time steps.
\subsection{Model training and model deployment}
In a scenario analogous to medical image segmentation, this study tackles a binary classification task on single-channel water occurrence probability images. \cite{Unet} employed the U-Net architecture for model training and deployment. The U-Net architecture comprises an encoder-decoder structure, where the encoder phase progressively extracts high-level features through convolutional operations, resulting in feature maps with a broader receptive field and a focus on global characteristics. During the decoding phase, deconvolutions are employed for upsampling, restoring the feature maps to the original input image size. In contrast to Fully Convolutional Networks\cite{FCN}, U-Net incorporates skip connections in the decoding phase to recover edge features that may have been lost during downsampling. This U-shaped architecture, addressing both high-level semantics and low-level features, enables U-Net to deliver promising outcomes in segmentation tasks, particularly in domains like medical imaging.

However, satellite high-resolution imagery possesses excessively large image dimensions, and the entities to be segmented are relatively sizeable. Traditional U-Net, constrained by the receptive field of convolutions, cannot effectively learn global and long-range semantic information interactions, exhibiting limitations in distinguishing between large rivers and large lakes in practical applications. To address this challenge, we employ the Swin-Unet architecture on our constructed training set\cite{Swin-Unet}. Swin-Unet substitutes all convolutions in U-Net with Transformers. In contrast to convolutional layers, whose receptive field is constrained by the kernel size, the Swin Transformer Blocks employed in Swin-Unet enable modeling global interactions among all pixels within the feature maps, circumventing the limitations imposed by the localized receptive fields of convolutions\cite{Swin}.

The architecture of Swin-Unet, as depicted in figure \ref{fig1}, commences with an input image partitioned into non-overlapping patches through a patch partitioning module akin to ViT's patch splitting\cite{ViT}. Each flattened patch is treated as a "token," and a linear embedding layer projects these raw feature tokens to an arbitrary dimension. The transformed patch tokens then propagate through several Swin Transformer blocks and patch merging layers to generate hierarchical feature representations. Specifically, the patch merging layers are tasked with downsampling and increasing the feature dimension, while the Swin Transformer blocks perform feature representation learning. The decoder, comprising Swin Transformer blocks and patch expanding layers, integrates the contextual features extracted from the encoder with multi-scale features via skip connections to compensate for the spatial information loss induced by downsampling. Analogous to the inverse process of patch merging layers, patch expanding layers effectuate upsampling by reshaping feature maps from adjacent dimensions into larger feature maps with double the resolution. Ultimately, the final patch expanding layer upsamples to restore the feature map resolution to the input resolution, after which a linear projection layer outputs pixel-level segmentation predictions on these upsampled features.

\begin{figure*}[!t]
\centering
\subfloat[]{\includegraphics[width=2.5in]{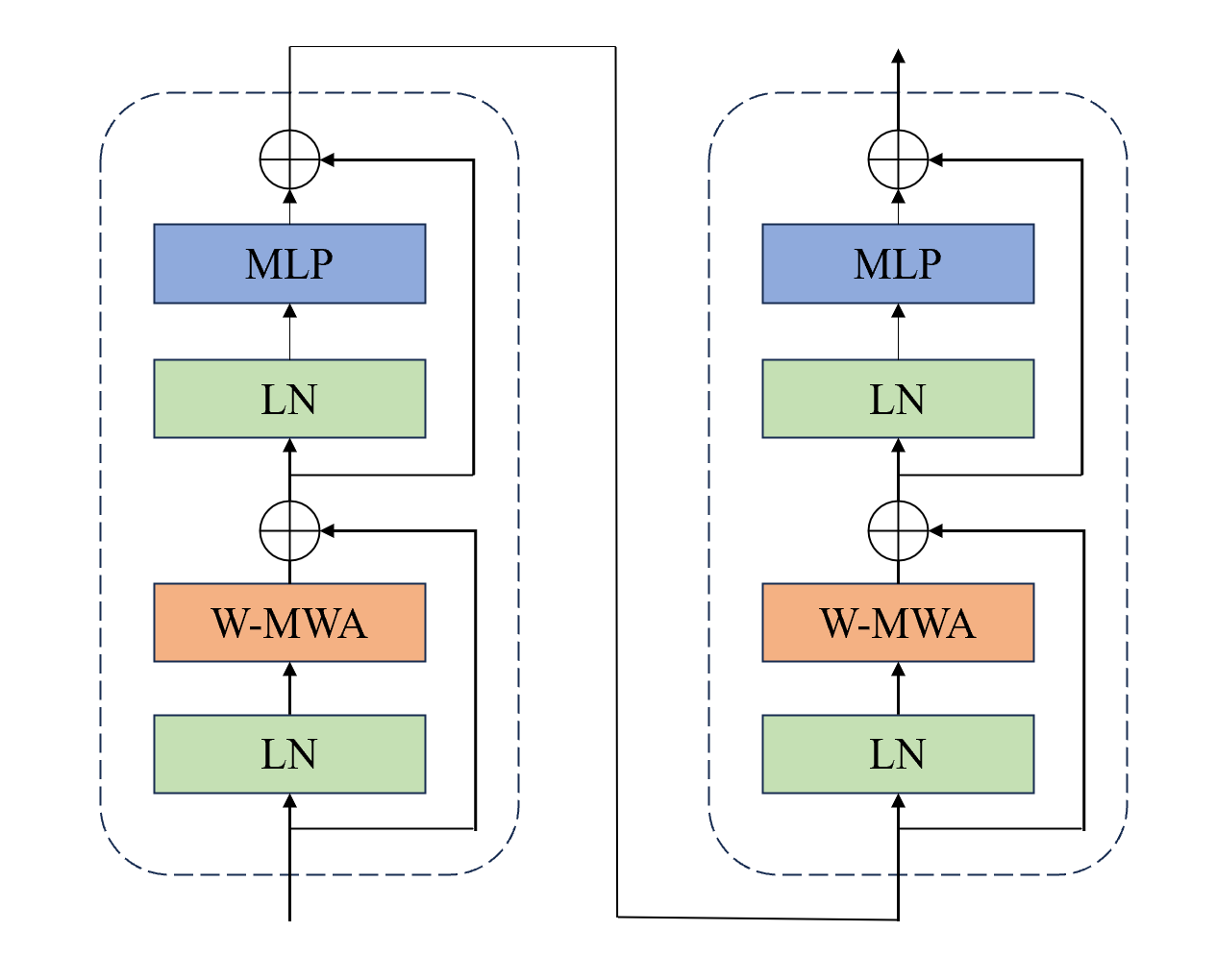}%
\label{fig2_a}}
\hfil
\subfloat[]{
\includegraphics[width=2.5in]{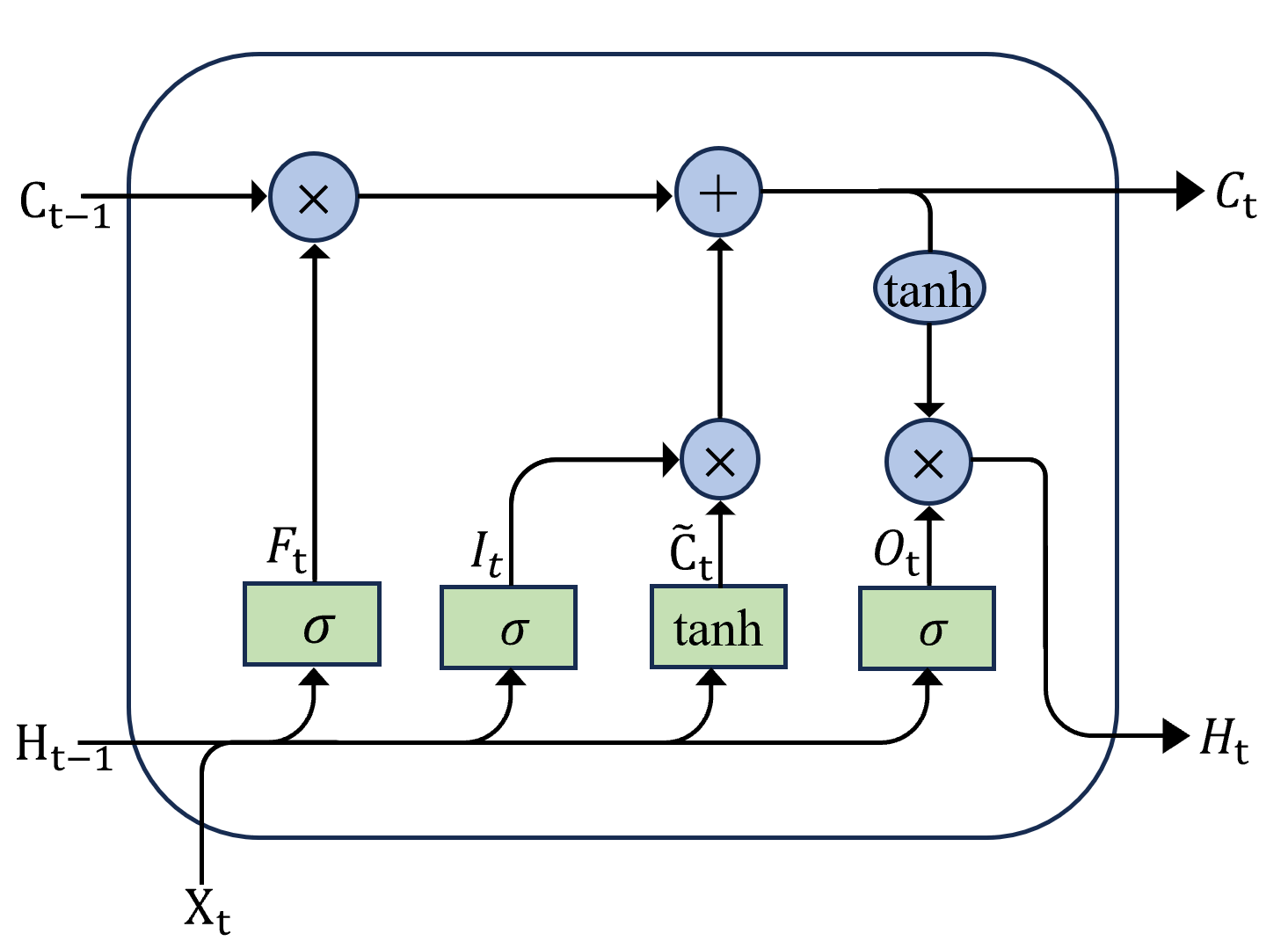}%
\label{fig2_b}}
\caption{ (a) Detailed structure of the Swin Transformer Block. (b) Architecture of the Long Short-Term Memory (LSTM) network.}
\label{fig2}
\end{figure*}

The architecture of the Swin Transformer Block is illustrated in Figure \ref{fig2_a}, comprising two consecutive multi-head self-attention modules and a 2-layer MLP with GELU activation\cite{GELU}. The preceding W-MSA module employs a windowing scheme that partitions the feature map into non-overlapping local windows, thereby restricting self-attention computation within each window to enhance computational efficiency, resulting in linear computational complexity with respect to image size. The subsequent SW-MSA module facilitates cross-window connections by shifting the window partitioning, enabling information exchange across windows. A LayerNorm layer\cite{LN} is applied before each multi-head attention module and MLP, followed by a residual connection after each multi-head attention module and MLP. Multiple Swin Transformer Blocks can be stacked, endowing this hierarchical architecture with the flexibility to model at various scales. The successive Swin Transformer Blocks can be mathematically formulated as:

\begin{equation}
\begin{aligned}
&\mathbf{\hat z}^l=W\mbox - MSA(LN(\mathbf z^l))+\mathbf z^{l-1}\label{5eq}\\
&\mathbf z^l=MLP(LN(\mathbf{\hat z}^l))+\mathbf{\hat z}^l\\
&\mathbf{\hat z}^{l+1}=SW\mbox - MSA(LN(\mathbf z^l))+\mathbf z^l\\
&\mathbf z^{l+1}=MLP(LN(\mathbf{\hat z}^{l+1}))+\mathbf{\hat z}^{l+1}
\end{aligned}
\end{equation}

In the formulation, the symbols $\mathbf{\hat z}^l$ and $\mathbf z^l$ denote the outputs from the SW-MSA module and the MLP module, respectively.And the self-attention mechanism is formulated as follows:
\begin{equation}
Attention(Q,K,V)=SoftMax(\frac{QK^T}{\sqrt{d}}+B)
\end{equation}
In the equation, $Q,K,V\in R^{M^2\times d}$represent the query, key, and value matrices, respectively, where $M^2$ denotes the number of patches within a window, and $d$ denotes the dimension of the query or key vector. The values in $B$ are derived from the bias matrix  $\hat B\in R^{(2M-1)\times(2N-1)}$.

To address the challenge posed by lakes situated in flood-prone regions exhibiting distinct frequency characteristics in water bodies compared to other areas, we adopted a similar approach to \cite{GLAKES} by segregating the flood-prone plains from other regions during the training process. Specifically, we calculated the proportion of pixels with water frequency less than 75\% within the river buffer zone. If this proportion exceeded a threshold of 0.1, the region was categorized as flood-prone, and a specialized Unet model was employed. For inference, we extracted regions of $512 \times 512$ pixels and utilized the Global Self-consistent Hierarchical High-resolution Shorelines (GSHHS) coastal dataset to exclude oceanic areas. Once a batch of regions was collected, it was transferred to the GPU for prediction, and the complete lake mask was obtained by merging the individual predictions.

\subsection{Contour extraction and data post-processing}
In this processing stage, we leverage the capabilities of the OpenCV library to detect the contours of lakes on a binary mask representing their presence. Contours can be conceptualized as curves that connect all continuous points along the boundary of an object. We specifically retain only the outermost contours to represent the lakes, while eliminating sub-contours that may arise, particularly in cases where lakes contain islands within their boundaries. 

To generate accurate GLAKES coverage without errors, \cite{GRWL} employed a manual approach, excluding lakes that overlapped more than 0.8 with river masks from the GRWL dataset. However, given the need to repeatedly process 16 sets of global data, a fully manual approach was impractical for our study. Consequently, we adopted a simpler and more robust method. We utilized the Simplified Vector Product derived from the GRWL masking products. This product is a smaller and more wieldy version of the raw GRWL vector dataset, making it more suitable for our large-scale processing needs. The Simplified Vector Product reduces the number of feature vertices and attributes by simplifying the polyline geometry and by calculating summary statistics along each polyline segment, where polyline segments are roughly the line segments between each tributary junction. From this data, we filtered rivers exceeding 1 km in width and obtained vector line segments representing the global distribution of these large rivers. Subsequently, we identified and discarded any lake contours that intersected with these vector segments, reducing the occurrence of false positives by eliminating water bodies mistakenly classified as lakes.

Furthermore, to establish a consistent and standardized indexing system for the detected lakes, we compare the identified water bodies against the GLAKES. During this comparison process, an R-tree spatial index is generated within the GLAKES dataset to enhance the efficiency of the search procedure. The R-tree is a tree data structure designed for indexing spatial data, allowing for efficient retrieval of objects based on their spatial relationships. Specifically, we first find all the lakes in GLAKES that intersect with each lake contour. For each contour, we select the GLAKES lake with the largest area of intersection as representing the same water body. If the maximum intersection area does not reach 30\% of the lake contour's area, we discard the contour as invalid data.

By integrating the contour detection approach, the GRWL dataset for river identification, and the GLAKES database for consistent indexing, we perform this operation on the sixteen sets of lake contour data from different time periods. This results in a dynamic vector dataset of lakes with some temporal gaps, where the time step between consecutive datasets is two years. Subsequently, we employ bidirectional linear interpolation to fill in the missing data gaps. During the interpolation process, we allow a maximum of three consecutive linear interpolations to ensure the quality and accuracy of the interpolated data. After completing these steps, we obtain a supplementary dataset to GLAKES, named CLAKES-Additional, which describes lake changes with high temporal resolution.

\subsection{Time series prediction}
To predict lake areas using the constructed lake dataset, we employed a stacked Long Short-Term Memory (LSTM) network to model the constructed CLAKES-Additional dataset.
LSTM is an improved variant of recurrent neural networks (RNNs) that addresses the issue of vanishing gradients, which can lead to difficulties in capturing long-term dependencies in sequential data. LSTM incorporate gate mechanisms that allow them to effectively combine short-term and long-term memory, enabling the preservation and propagation of relevant information across long sequences. 

To better represent the process, Let us consider a time series dataset comprising $M$ sequences, each with a length of $N$ time steps. We denote the j-th sequence as $\{\mathbf x_1^j,\mathbf x_2^j,\mathbf x_3^j,\dots,\mathbf x_N^j\}$, where $j=1,2,3,\dots,M$, and $\mathbf x_t^j$ is a vector representing lake area, precipitation, vapor pressure, and temperature at time step $t$ for the $j$-th sequence. To combine all the $\mathbf x_1$ vectors into a matrix $\mathbf X_1$, we construct ${\mathbf X_1 = [\mathbf x_1^1,\mathbf x_1^2,\mathbf x_1^3,\dots,\mathbf x_1^M]^T}$

The initial step of the forward propagation procedure in the Long Short-Term Memory (LSTM) network involves computing the gate activations and candidate cell state. Mathematically, this can be formulated as:
\begin{equation}
\begin{aligned}
\mathbf{F}_t &= \sigma([\mathbf{X}_t,\mathbf{H}_{t-1}] \mathbf{W}_{f} + \mathbf{b}_f^T)\\
\mathbf{I}_t &= \sigma([\mathbf{X}_t,\mathbf{H}_{t-1}]\mathbf{W}_{i} + \mathbf{b}_i^T)\\
\mathbf{O}_t &= \sigma([\mathbf{X}_t,\mathbf{H}_{t-1}] \mathbf{W}_{o} + \mathbf{b}_o^T)\\
\tilde{\mathbf{C}}_t &= \text{tanh}([\mathbf{X}_t,\mathbf{H}_{t-1}]\mathbf{W}_{c} + \mathbf{b}_c^T),
\end{aligned}
\end{equation}

Where $\mathbf{F}_t$, $\mathbf I_t$, and $\mathbf O_t$ represent the forget, input, and output gate activations, respectively, at time step $t$. $\tilde{\mathbf{C}}_t$ denotes the candidate cell state. $\sigma$ and $\text{tanh}$ are the sigmoid and hyperbolic tangent activation functions, respectively.$\mathbf W_f$,$\mathbf W_i$, $\mathbf W_o$, and $\mathbf W_c$ are the weight matrices associated with the respective gates and the candidate cell state, while $\mathbf b_f$, $\mathbf b_i$, $\mathbf b_o$, and $\mathbf b_c$ are the corresponding bias vectors. The inputs to these computations are the current input $\mathbf X_t$ and the previous hidden state $\mathbf H_1$, concatenated together.

Subsequently, the new memory cell $\mathbf C_t$ is computed as a gated combination of the previous cell state $\mathbf C_{t-1}$ and the candidate cell state $\tilde{\mathbf{C}}_t$, modulated by the forget gate $\mathbf F_t$ and input gate $\mathbf I_t$, respectively.And the hidden state $\mathbf H_t$ is derived from the current memory cell $\mathbf C_t$, regulated by the output gate $\mathbf O_t$::
\begin{equation}
\begin{aligned}
\mathbf{C}_t &= \mathbf{F}_t \odot \mathbf{C}_{t-1} + \mathbf{I}_t \odot \tilde{\mathbf{C}}_t\\
\mathbf{H}_t &= \mathbf{O}_t \odot \tanh(\mathbf{C}_t)
\end{aligned}
\end{equation}
Where $\odot$ denotes element-wise multiplication, and tanh is the hyperbolic tangent activation function. This formulation enables the LSTM to selectively retain and propagate relevant information through the sequence, capturing long-term dependencies while generating context-aware hidden representations.

Following the generation of the hidden state representations, an output function is employed to map the hidden states to the desired output space. The discrepancy between the predicted outputs and the corresponding ground truth values is then quantified using the Mean Squared Error (MSE) loss function:
\begin{equation}
L(\hat y,y)=\frac{1}{M}\sum_{j=1}^{M}(\hat y-y)^2
\end{equation}
By minimizing the MSE through training, we can optimize the model parameters to make the predictions closer to the true values.
\section{Experimental results and analysis}
\subsection{Evaluation metrics}
Semantic segmentation is a pixel-level classification task, and the commonly employed evaluation metrics include Pixel Accuracy (PA) and Mean Intersection over Union (MIoU). PA represents the proportion of correctly classified pixels among the total number of pixels, while MIoU is a more comprehensive measure that calculates the ratio of the intersection and union of the predicted and ground truth segmentation masks for each class, and then averages these ratios across all classes. For binary classification problems, the specific formulations are as follows:

\begin{equation}
    P A=\frac{T P+T N}{T P+T N+F P+F N}
    \end{equation}

\begin{equation}
        M I o U=\frac{1}{2}\left(\frac{T P}{T P+F P+F N}+\frac{T N}{T N+F P+F N}\right)
\end{equation}

\subsection{Implementation details}
The Swin-Unet is implemented using the PyTorch framework on a system equipped with an Intel(R) Xeon(R) Platinum 8352V CPU @ 2.10GHz, 256GB memory, and an NVIDIA RTX A6000 GPU. 80\% of the prepared dataset is utilized for training, while the remaining 20\% serves as the test dataset. The Adam optimizer is employed, and the initial learning rate is set to 0.005. If the mean intersection over union (MIoU) on the validation set does not improve for 6 consecutive epochs, the learning rate is decayed to 0.33 times its previous value.

To address the issue of imbalanced sample distributions in the training set, the Tversky loss function is employed as the loss function to mitigate this problem. The Tversky loss function can be formulated as follows:
\begin{equation}
    T(\alpha, \beta)=\frac{\sum_{i=1}^N p_{0 i} g_{0 i}}{\sum_{i=1}^N p_{0 i} g_{0 i}+\alpha \sum_{i=1}^N p_{0 i} g_{1 i}+\beta \sum_{i=1}^N p_{1 i} g_{0 i}}
    \end{equation}

In the formulation, $p_{0i}$ and $p_{1i}$ represent the probabilities outputted by the softmax layer corresponding to the pixel $i$ being classified as a non-lake and lake, respectively. Additionally, $g_{0i}$and $g_{1i}$ denote the ground truth labels for pixel $i$, where $g_{0i}$ takes the value of 1 for a non-lake pixel and 0 for a lake pixel, while $g_{1i}$ represents the complement, being 0 for a non-lake voxel and 1 for a lake pixel. The hyperparameters $\alpha$ and $\beta$ enable the modulation of the trade-off between false positive and false negative errors. Their judicious adjustment allows calibrating the relative importance of precision versus recall, tailoring the loss function to the desired performance prioritization based on application requirements.

\begin{figure*}[htbp]
    
\centering
\subfloat[]{
\includegraphics[width=3.5in]{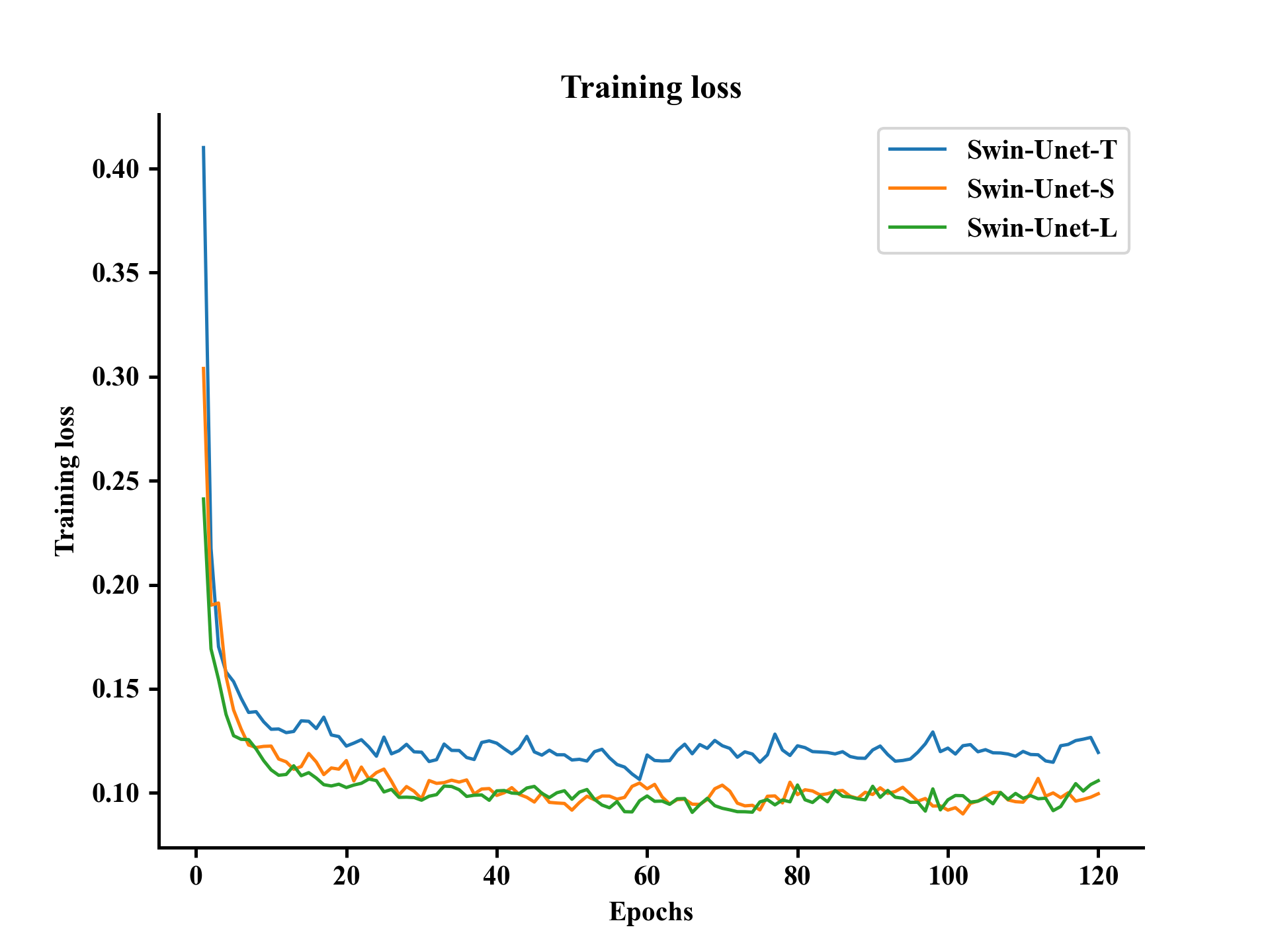}%
\label{fig4_a}}
\hfil
\subfloat[]{
\includegraphics[width=3.5in]{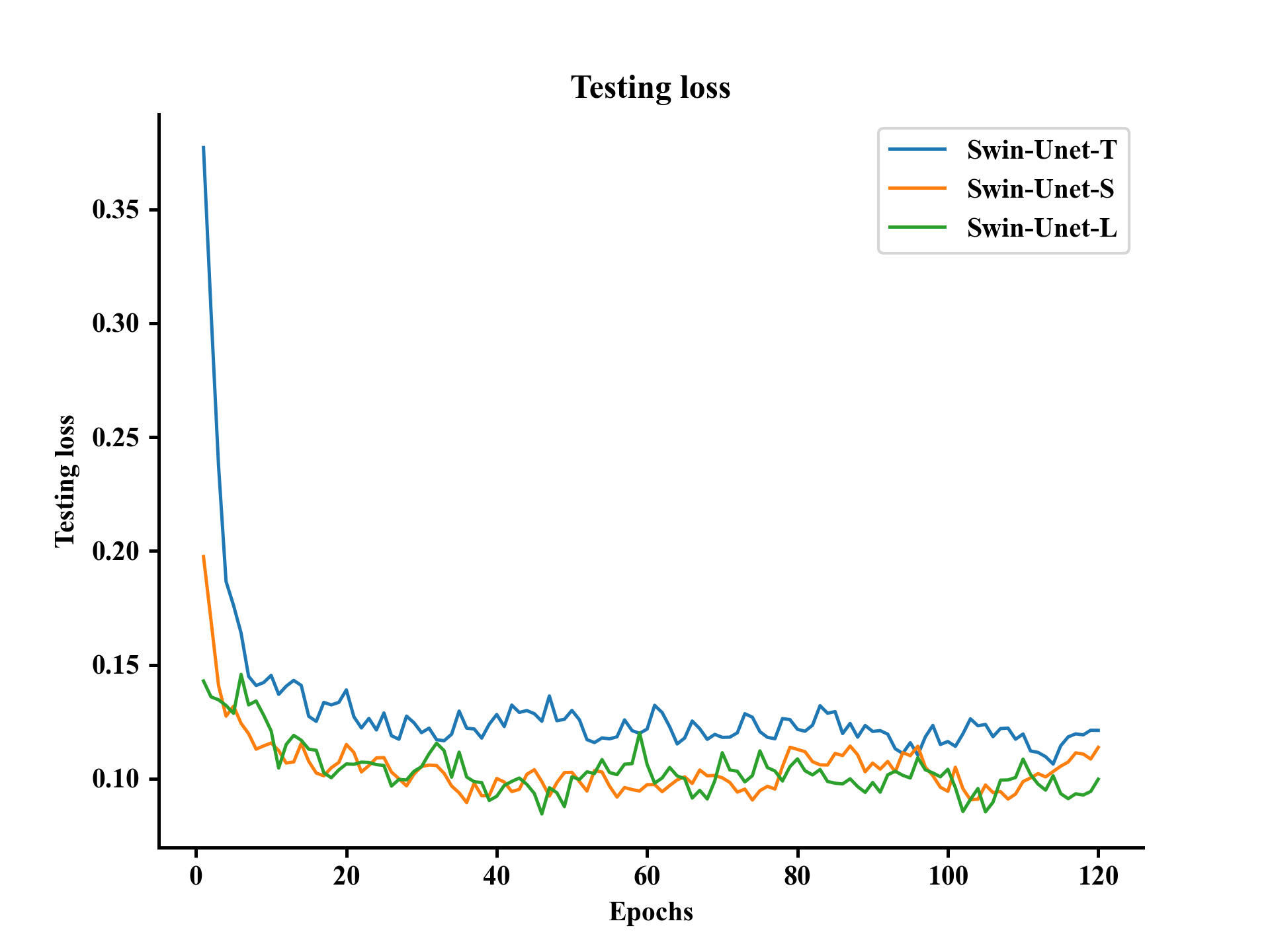}%
\label{fig4_b}}
\caption{ Loss curve during the (a) training and (b) testing process of Swin-Unet.}
\label{fig4}
\end{figure*}
\begin{figure*}[]
\centering
\subfloat[]{
\includegraphics[width=3.5in]{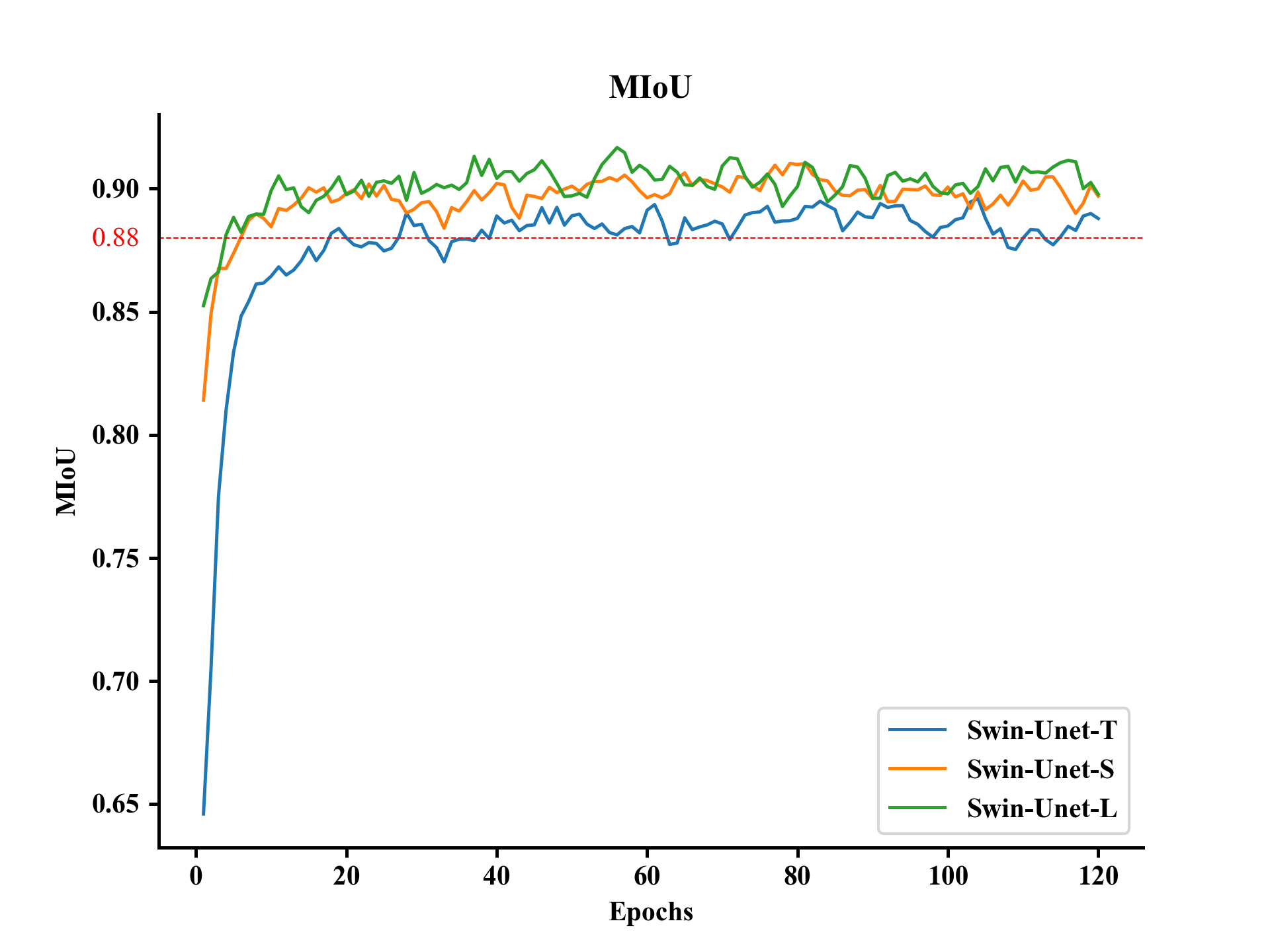}%
\label{fig5_a}}
\hfil
\subfloat[]{
\includegraphics[width=3.5in]{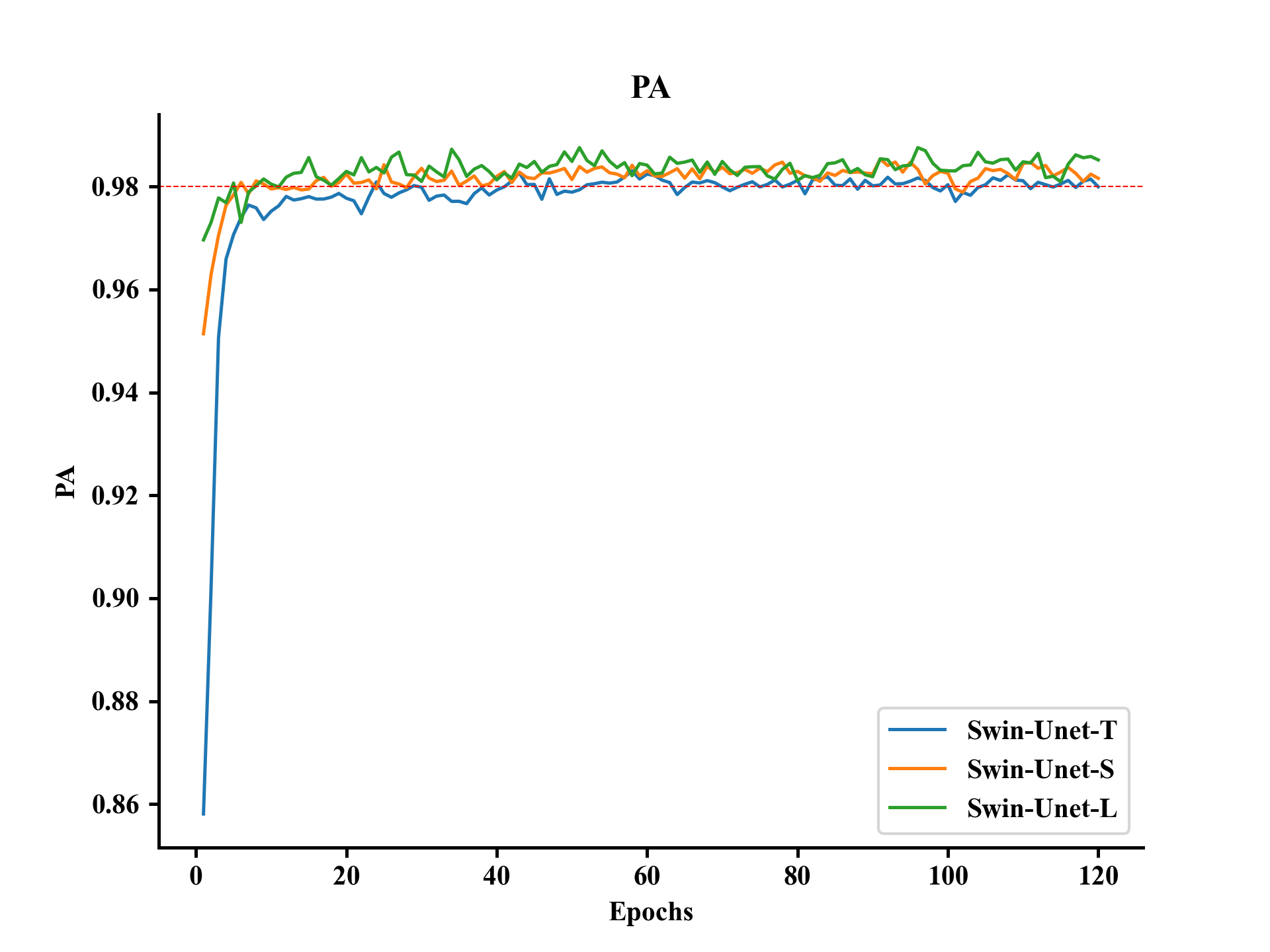}%
\label{fig5_b}}
\caption{Evaluation indexes of the semantic segmentation results. (a) Mean Intersection over Union (b) Pixel Accuracy.}
\label{fig5}
\end{figure*}
In our experiments, we extend the Swin-Unet architecture to 5 layers, effectively performing 4 patch merging operations. By adjusting the number of Swin Transformer blocks per layer and the number of attention heads in the multi-head attention modules, we train three variants of the Swin-Unet with different parameter configurations, as detailed in Table \ref{table1}. The trends of loss values during the training and testing processes for the different Swin-Unet variants represent the discrepancies between the predicted and true values, as illustrated in Fig. \ref{fig4}. In the initial 20 epochs, the loss function value drops sharply. The rapid decrease in loss during the training process indicates that the proposed Swin-Unet can localize the target positions with high convergence speed.

The evolution of the evaluation metrics, mean Intersection over Union (MIoU) and Pixel Accuracy (PA), with respect to the number of iterations during the training process is depicted in Figures 5a and 5b, respectively. It can be observed that the variant with the smallest parameter count, Swin-Unet-T, exhibits noticeable deficiencies compared to Swin-Unet-S and Swin-Unet-L. However, it still achieves and even slightly surpasses the final MIoU of 0.88 and PA of 0.98 attained by the U-Net architecture employed by \cite{GLAKES}. The Swin-Unet-S surpasses the baseline performance after just 10 iterations, ultimately converging to an MIoU of 0.905. Ultimately, while the Swin-Unet-L variant features an increased parameter count compared to Swin-Unet-S, its design and performance do not exhibit a significant improvement.

\begin{table*}[]\centering

\tabcolsep=1cm
\caption{Detailed architecture specifications\label{table1}}
\begin{tabular}{ccccc}
\toprule
model        & Blocks        &Heads              & patch size                 &window size\\
\midrule
Swin-Unet-T  & 2,4,6,8,10    &12,24,48,96,192    & \multirow{3}{*}{4$\times$4}&\multirow{3}{*}{7$\times$7}\\
Swin-Unet-S  & 2,2,4,6,8     &6,12,24,48,96      &                            &\\ 
Swin-Unet-L  & 2,2,2,4,6     &3,6,12,24,48       &                            &\\
\bottomrule
\end{tabular}
\end{table*}

\subsection{Comparisons with other methods}
Considering the application scenario of this study involves a single-channel binary semantic segmentation task with relatively simple image semantics, fixed structures, and limited data volume, similar to medical image segmentation applications, we conducted a comparative experimental analysis to evaluate the performance of Swin-UNet against U-Net and its variants, including Res-UNet, UNet++, and U2-Net, which have demonstrated promising results in the medical image segmentation domain. Specifically, Res-UNet incorporates residual connections in each encoder and decoder module to facilitate the extraction of low-frequency information and alleviate the vanishing gradient problem\cite{Res-Unet}, U-Net++ adopts the dense connection approach inspired by DenseNet, enabling enhanced feature reconstruction and retention of both global and local information\cite{Unet++}, and U2-Net increases the depth of the architecture by employing Residual U-blocks without significantly increasing the computational cost\cite{U2-net}.

As illustrated in Table \ref{table2}, using the evaluation metrics of Mean Intersection over Union (MIoU) and mean Pixel Accuracy (mPA), the results indicate that Swin-UNet outperformed U-Net and its variants in terms of both MIoU (0.915) and mPA (0.951) scores. Consequently, Swin-UNet demonstrated superior performance, meeting the expected requirements for the satellite image segmentation task in this application scenario.
\begin{table}[!t]
\centering
\tabcolsep=1cm
\caption{Comparison of evaluation indicators for different semantic segmentation methods.\label{table2}}
\begin{tabular}{ccc}
    \toprule
    model&MIoU& mPA\\
    \midrule
    U-Net & 88.1 & 92.5\\
    Res-UNet & 89.8 & 93.4\\
    UNet++ & 89.4 & 92.7\\
    U2-Net & 87.1 & 91.8\\
    Swin-UNet & 91.5 & 95.1\\
    \bottomrule
\end{tabular}

\end{table}

\begin{figure*}[!t]
\centering
\includegraphics[width=1\textwidth]{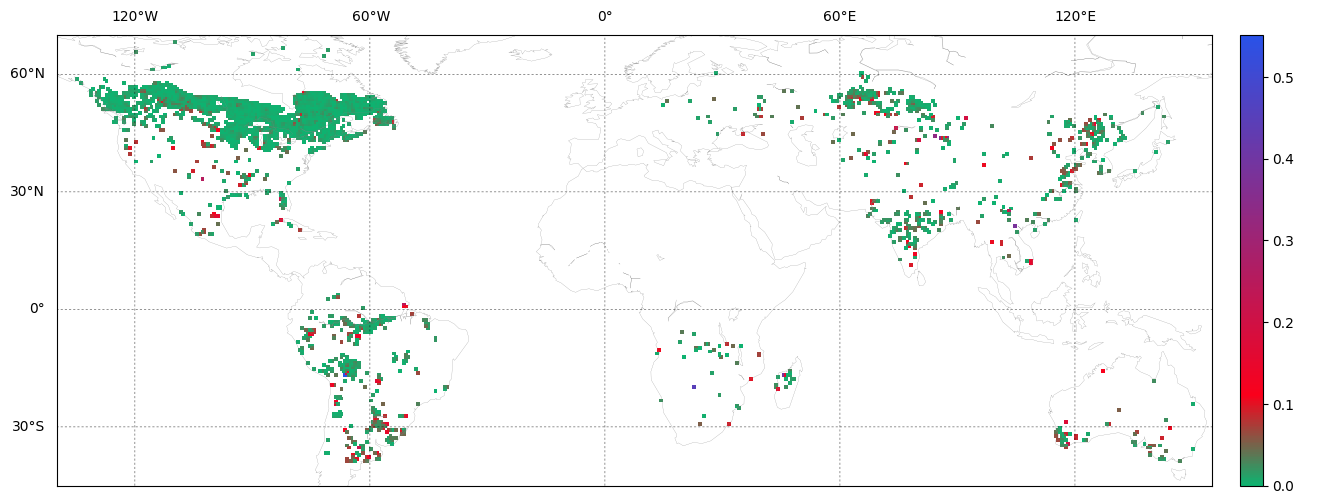}
\caption{Global map of lake area prediction errors from the LSTM model. The absolute values of prediction errors are normalized by the corresponding lake areas from the previous time step and plotted on a 0.5° longitude by 0.5° latitude grid.}
\label{Pred_Global}
\end{figure*}

\subsection{Applications to area prediction}
We retrieved meteorological data, including temperature, vapor pressure, and precipitation, for each lake corresponding to the years provided in the GLAKES-Additional dataset, from the CRU TS meteorological dataset. Lakes supplied by glaciers or permafrost, as well as those impacted by human activities, were excluded based on the lake type labels in the GLAKES dataset. This ensured that the lakes included in our analysis were directly influenced by climate change factors, eliminating potential confounding factors. Due to the heightened sensitivity of small and medium-sized lakes to climate change, we selected lakes ranging from 1 square kilometers to 60 square kilometers in surface area for our experimental study. Subsequently, the lakes were divided into 10 levels based on the average area over multiple years, to account for potential differences in lake behavior due to size variations. Employing a stratified sampling approach, we obtained balanced trainning sets and testing sets with respect to lake areas, ensuring that the datasets were representative of the entire range of lake sizes. To process the area and meteorological sequences of each lake in the training set, a sliding window methodology was applied. This approach involves dividing the time series data into overlapping windows, allowing the model to capture temporal patterns and dependencies within the data. By considering multiple time steps simultaneously, the model can learn to recognize and predict the evolving dynamics of lake area changes in response to meteorological conditions. Furthermore, data from multiple lakes were aggregated to describe common patterns of lake changes under the influence of climate change. 

\begin{figure}[!t]
\centering
\includegraphics[width=3.4in]{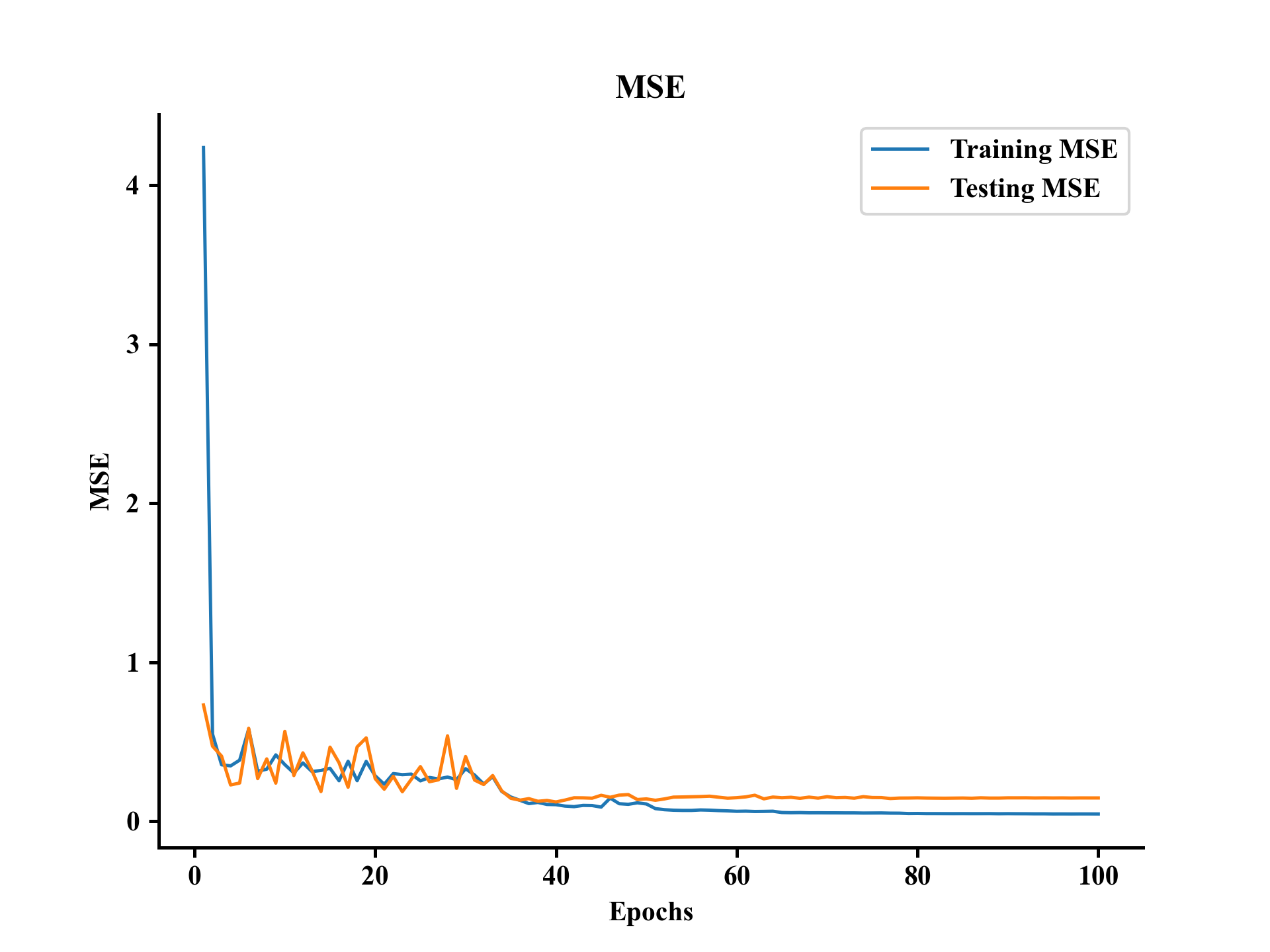}
\caption{MSE curves of the training and test sets for lake area time-series prediction using an LSTM model.}
\label{Fig_MSE}
\end{figure}

Subsequently, a two-layer LSTM model was trained, and the variation of MSE on the training and test sets with respect to the number of iterations is illustrated in Figure \ref{Fig_MSE}. During the first 5 epochs, the MSE decreased rapidly; however, after 18 epochs, the MSE tended to stabilize, albeit with gradual signs of overfitting. Ultimately, the model achieved a stable MSE of 0.101 and RMSE of 0.317 square kilometers on the test set. This result indicates that the lake area change information provided by GLAKES-Additional is sufficient to enable the LSTM model to learn the patterns of lake area variations under the influence of climate change.

To demonstrate the global performance of the trained LSTM model across different regions, we plotted the absolute values of the lake area prediction errors, normalized by the corresponding lake areas from the previous time step, on a global map with a 0.5° longitude by 0.5° latitude grid resolution, as shown in Fig. \ref{Pred_Global}. From this figure, it can be observed that the LSTM model performs relatively well in areas with high lake density, particularly evident in the North American Great Lakes region. This is attributed to the availability of more samples in densely populated lake areas, which supports the LSTM model in learning the patterns of lake area changes influenced by climate change within those regions. The North American Great Lakes region, harboring a large number of lakes and benefiting from good Landsat satellite image coverage, contains the highest concentration of lakes in the GLAKES-Additional dataset, as illustrated in Fig. \ref{Fig1}. Additionally, there are a few outliers where the error exceeds 0.3 of the previous time step's lake area, suggesting that the lake area changes in these cases are more significantly influenced by human activities and other factors. 

\begin{figure}[!t]
\centering
\includegraphics[width=3.4in]{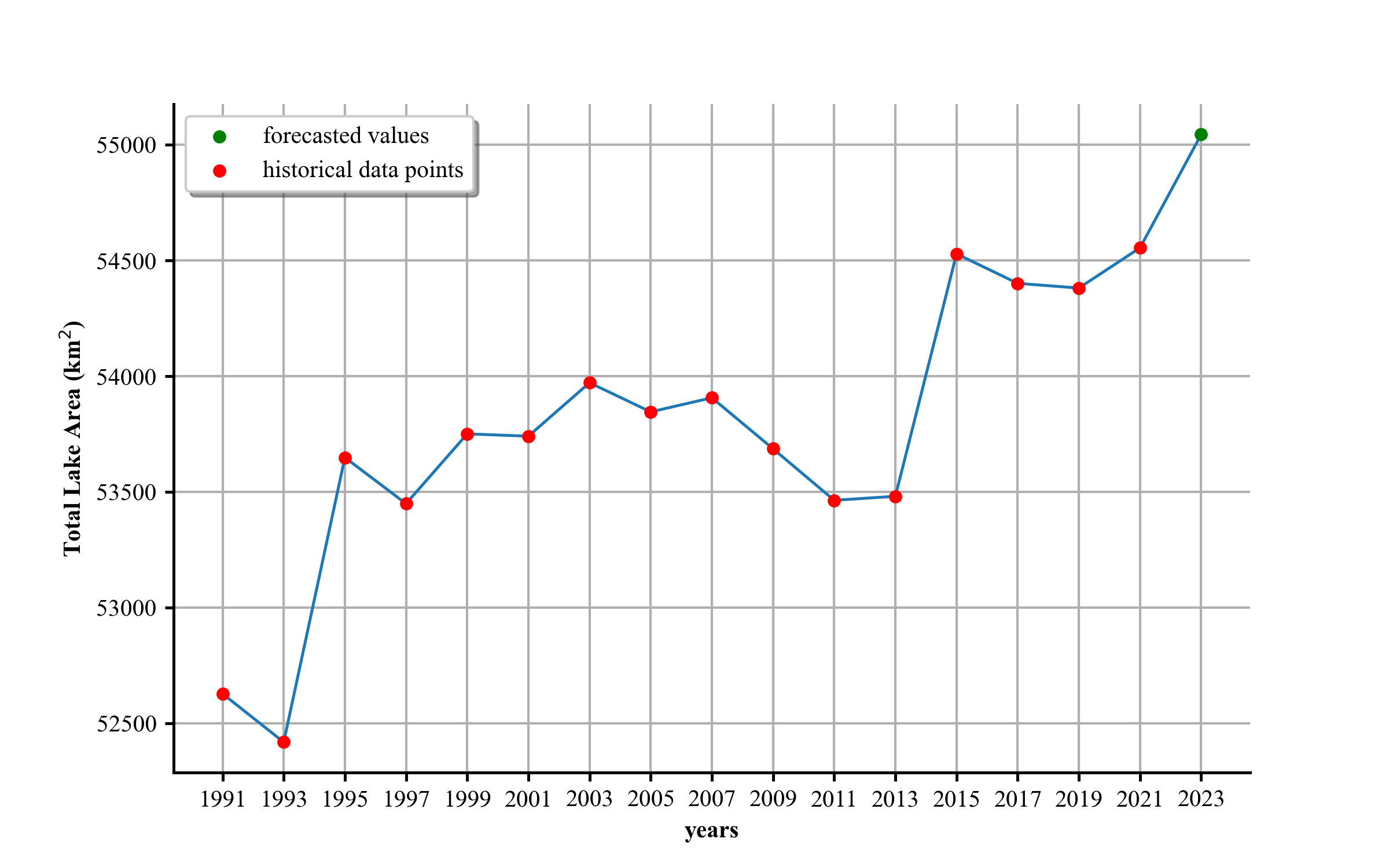}
\caption{Temporal evolution of the cumulative surface area of climate-sensitive lakes with an area between 1$km^2$ and 60$km^2$ in the GLAKES-Additional dataset from 1991 to 2023. The data points from 1991 to 2021 represent historical records, while the data point for 2023 is the predicted value obtained from the trained LSTM model.}
\label{Area_Sum}
\end{figure}

\begin{figure*}[!t]
\centering
\includegraphics[width=1\textwidth]{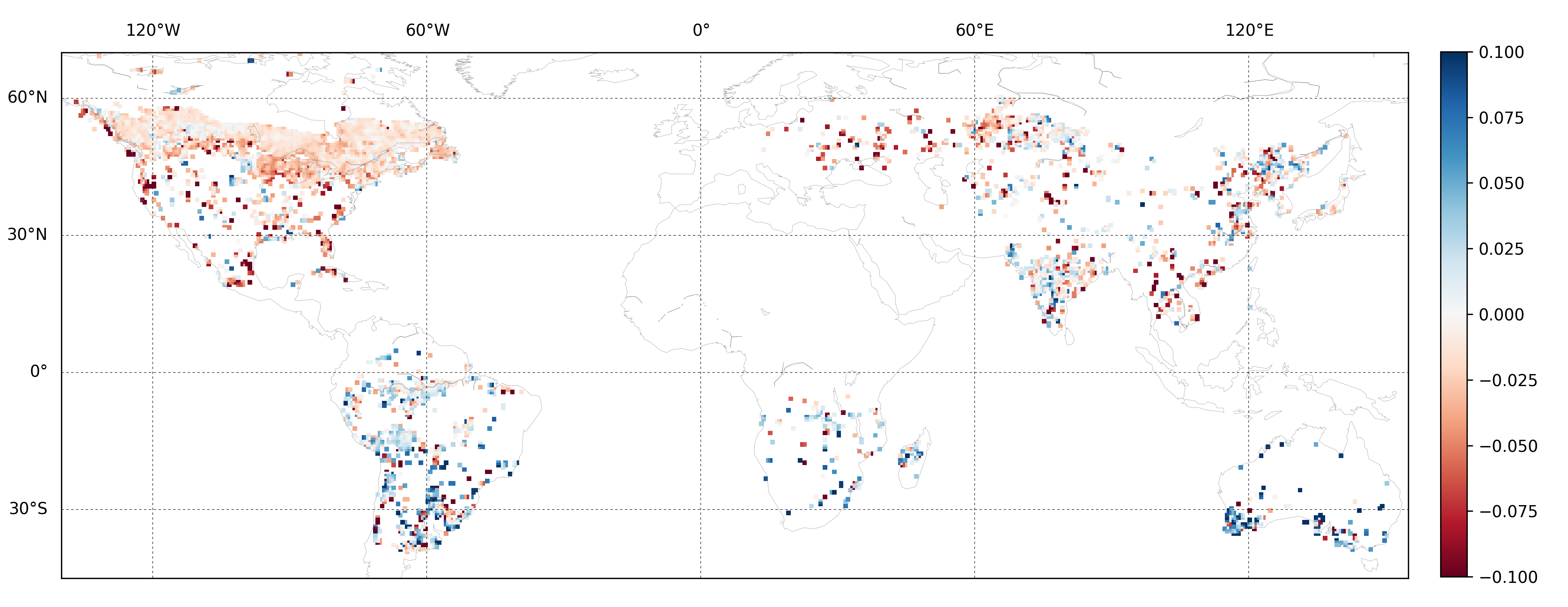}
\caption{Global distribution of the relative change in surface area between 2023 (predicted values) and 2021 (historical records) for climate-sensitive lakes with an area between 1$km^2$ and 60$km^2$ in the GLAKES-Additional dataset. The globe is divided into 0.5° longitude × 0.5° latitude grid cells, and each cell represents the relative change in the cumulative surface area of the lakes within that grid cell.}
\label{Pred_2023}
\end{figure*}

Subsequently, the trained LSTM model was utilized to predict the surface area data for all climate-sensitive lakes ranging from 1$km^2$ to 60$km^2$ in the GLAKES-Additional dataset for the year 2023. The cumulative surface area amounted to 55,044.28$km^2$, representing an increase of 489.67$km^2$ compared to the total lake surface area in 2021. The historical records and projected trends of lake surface area from 1991 to 2023 are depicted in Fig. \ref{Area_Sum}, exhibiting an overall upward trajectory. The spatial distribution of the relative change in lake surface area between 2023 and 2021 is presented in Fig. \ref{Pred_2023}, highlighting pronounced regional variations. North America experienced a predominant decline in lake surface area, whereas South America, southern Africa, and Oceania exhibited a general increase in lake surface area.

\section{Conclusion and future work}
Leveraging the GSW dataset, this study employed the Swin-UNet architecture to extract, at the highest possible temporal resolution, 152,567 lakes spanning from 1990 to 2021, thereby constructing the GLAKES-Additional dataset that characterizes the dynamic changes in lake extent. The efficacy of the Swin Transformer Block in addressing the limited receptive field issue faced by traditional convolutional semantic segmentation networks when applied to high-resolution satellite imagery is demonstrated. Furthermore, we utilized the lake area data from GLAKES-Additional, in conjunction with meteorological data, to perform time-series forecasting, validating the effectiveness of GLAKES-Additional.

Owing to a certain degree of data missingness in the MWH dataset, the minimum suitable time step that could be selected was limited to two years, resulting in only 16 time steps for generating GLAKES-Additional, which is somewhat inadequate for time-series forecasting. Additionally, the two-year interval rendered the area time series devoid of seasonality. To achieve higher temporal resolution, training models on more abundant raw satellite imagery should be considered for extraction purposes.
\section*{Acknowledgments}
This research was financially supported by the National Natural Science
Foundation of China (No. 42276203, 42030406), the Natural Science
Foundation of Shandong Province (No. ZR2021MD001), and the Laoshan
Laboratory (No. LSKJ202204302). 
\bibliographystyle{IEEEtran}
\bibliography{references}
\end{document}